\documentclass[runningheads]{llncs}
\usepackage[T1]{fontenc}
\usepackage{graphicx,verbatim}
\usepackage{amsmath,amssymb}
\usepackage{bbm}
\usepackage{multirow}
\usepackage{bbding}
\usepackage{color}
\usepackage{booktabs}
\usepackage{url}
\usepackage {arydshln}
\usepackage{multirow}
\usepackage[table]{xcolor}
\usepackage{bbding}
\usepackage{pifont}

\usepackage{marvosym}
\usepackage[colorlinks=true,linkcolor=blue,citecolor=blue]{hyperref}
\usepackage{arydshln}
\usepackage{threeparttable}
\usepackage{caption}

\begin{document}
%
\iffalse
\def\etal{\textit{et al}. }
\def\ie{\textit{i.e.}}
\def\eg{\textit{e.g.}}
\def\vs{\textit{v.s.}}
\fi

\def\etal{{et al}.}
\def\ie{{i.e.}}
\def\eg{{e.g.}}
\def\vs{{v.s.}}

% \title{Optimal Scoring Policy Modulation with Meta-Learner for Few-Shot Whole Slide Image Classification}
% \title{Meta-Optimized Scoring Policy Modulation for Few-Shot Whole Slide Image Classification}
% \title{Learn to Learn for Few-Shot Whole Slide Image Classification}
% learn an optimal classifier scheme for xx
% meta-optimized classifier scheme for xxx
% A meta-optimized adaptive classifier for few-shot
% \title{Adaptive Meta-Classifier for Robust Few-Shot Whole Slide Image Classification}
\title{MOC: Meta-Optimized Classifier for Few-Shot Whole Slide Image Classification}
\titlerunning{Meta-Optimized Classifier for Few-Shot WSI Classification}
%
\begin{comment}  %% Removed for anonymized MICCAI 2025 submission
\author{First Author\inst{1}\orcidID{0000-1111-2222-3333} \and
Second Author\inst{2,3}\orcidID{1111-2222-3333-4444} \and
Third Author\inst{3}\orcidID{2222--3333-4444-5555}}
%
\authorrunning{F. Author et al.}
\institute{Princeton University, Princeton NJ 08544, USA \and
Springer Heidelberg, Tiergartenstr. 17, 69121 Heidelberg, Germany
\email{lncs@springer.com}\\
\url{http://www.springer.com/gp/computer-science/lncs} \and
ABC Institute, Rupert-Karls-University Heidelberg, Heidelberg, Germany\\
\email{\{abc,lncs\}@uni-heidelberg.de}}

\end{comment}

% \author{Anonymized Authors}  %% Added for anonymized MICCAI 2025 submission
% \authorrunning{Anonymized Author et al.}
% \institute{Anonymized Affiliations \\
%     \email{email@anonymized.com}}

\author{Tianqi Xiang\inst{1} \and
Yi Li\inst{1} \and Qixiang Zhang\inst{1} \and Xiaomeng Li\inst{1,\thanks{Corresponding Author}}}

\institute{Department of Electronic and Computer Engineering, The Hong Kong University of Science and Technology, Hong Kong SAR, China\\ \email{eexmli@ust.hk}
}

\maketitle              % typeset the header of the contribution

\begin{abstract}
Recent advances in histopathology vision-language foundation models (VLFMs) have shown promise in addressing data scarcity for whole slide image (WSI) classification via zero-shot adaptation. However, these methods remain outperformed by conventional multiple instance learning (MIL) approaches trained on large datasets, motivating recent efforts to enhance VLFM-based WSI classification through few-shot learning paradigms. While existing few-shot methods improve diagnostic accuracy with limited annotations, their reliance on conventional classifier designs introduces critical vulnerabilities to data scarcity. To address this problem, we propose a Meta-Optimized Classifier (MOC) comprising two core components: (1) a meta-learner that automatically optimizes a classifier configuration from a mixture of candidate classifiers and (2) a classifier bank housing diverse candidate classifiers to enable a holistic pathological interpretation. Extensive experiments demonstrate that MOC outperforms prior arts in multiple few-shot benchmarks. Notably, on the TCGA-NSCLC benchmark, MOC improves AUC by 10.4\% over the state-of-the-art few-shot VLFM-based methods, with gains up to 26.25\% under 1-shot conditions, offering a critical advancement for clinical deployments where diagnostic training data is severely limited. Code is available at \url{https://github.com/xmed-lab/MOC}.
\end{abstract}

\section{Introduction}
% Vision-language models for pathology images xxx, CLIP, CONCH. However, when performing zero-shot evaluation on xxxx, limited results. This trigger researchers to propose few-shot WSI for VLM which requires a limited labeled samples (e.g., 1, 4 ) in order to generalize the foundations models to xxxx. 

% Most of existing few-shot methods for VLM pathology aims to xxx prompt-tuning xxx to fine-tune xxxx [xxx,xx cite papers]. For example, TOP. Based on TOP, MCPAT. FAST. However, these methods mainly focus on xxx, without considering the aggregation. xxxx. We found that xxx agrregatiuon xxx. 

% Recent pathology vision-language foundation models (FMs) like PLIP~\cite{PLIP}, BiomedCLIP~\cite{BiomedCLIP}, and CONCH~\cite{CONCH} demonstrate strong histopathology understanding and open-world generalization via vocabulary. \textcolor{red}{Unlike conventional multiple instance learning (MIL) frameworks that rely on large annotated datasets, these FMs enable zero-shot adaptation through visual-language alignment, addressing data scarcity due to labor, privacy, or rare diseases.}(先说FM利用的是zero shot，有什么样的好处。在对比MIL，强调有gap。所以考虑用few shot，取一个trade off，实现xxx). However, a performance gap remains between FMs and fully supervised MIL methods, stimulating research into few-shot fm WSI classification that enhances diagnostic accuracy with limited labeled histopathology slides.

Recent advances in pathology vision-language foundation models (VLFMs), such as PLIP~\cite{PLIP}, BiomedCLIP~\cite{BiomedCLIP}, and CONCH~\cite{CONCH}, have demonstrated remarkable capabilities in histopathological image interpretation through visual-language alignment. These models mitigate data scarcity challenges stemming from annotation costs, privacy constraints, and rare disease prevalence via zero-shot adaptation. However, existing VLFM-based zero-shot approaches exhibit inferior performance compared to conventional multiple instance learning (MIL) frameworks that leverage extensive annotated whole slide image (WSI) datasets. This performance discrepancy has motivated emerging research into few-shot VLFM-based WSI classifications that enhance diagnostic reliability with minimal annotated data.

\begin{figure}[t]
\centering
\resizebox{1\textwidth}{!}{
\includegraphics[width=\textwidth]{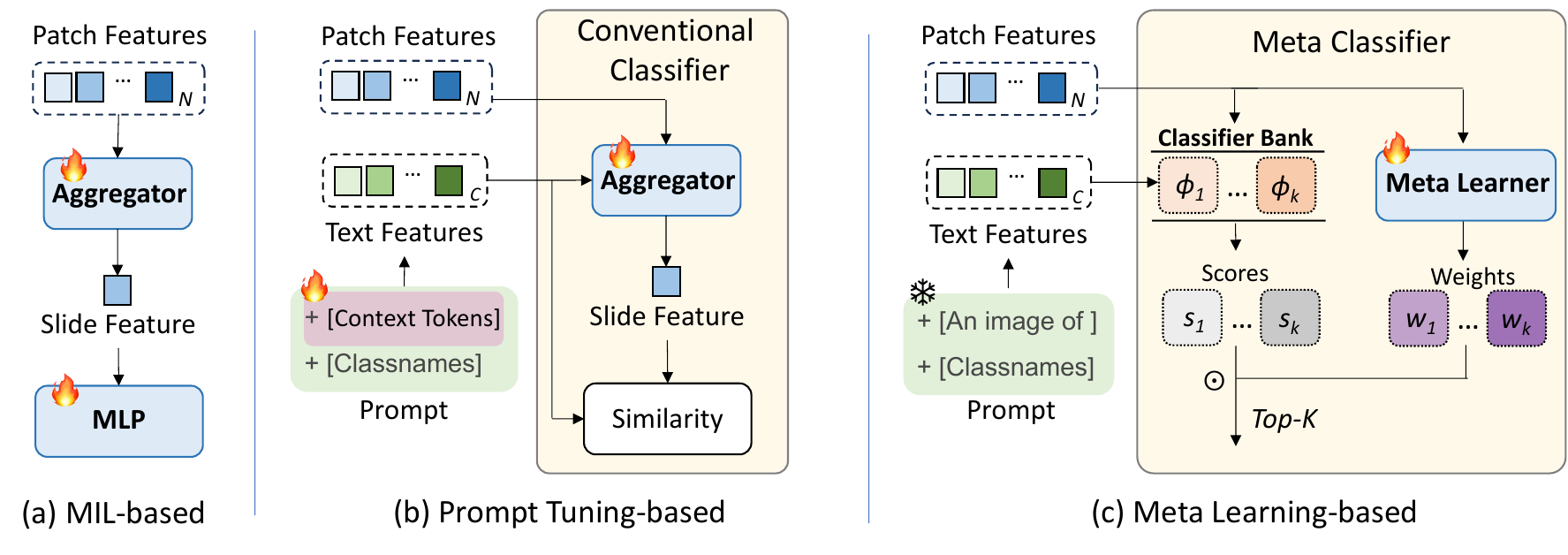}}
% \vspace{-7mm}
\caption{Architectural comparison among (a) MIL-based methods, (b) prompt tuning-based methods, and (c) our proposed meta learning-based method for VLFM few-shot pathology analysis. Different from prompt-tuning-based frameworks that adopt \textbf{conventional classifiers}, our method uses the meta learner to dynamically compose an optimal \textbf{meta classifier} from the classifier bank.}
\label{intro_compare}
\end{figure}

Existing VLFM-based few-shot WSI classification methods (Fig.~\ref{intro_compare}(b)), distinguish themselves from conventional MIL frameworks~\cite{ABMIL,CLAM,transmil,dsmil,dtfdmil,gtmil,ibmil,mhimmil} (Fig.~\ref{intro_compare}(a)) by incorporating linguistic supervision alongside visual understanding.
% Existing VLFM-based few-shot WSI classification methods, as illustrated in Fig.~\ref{intro_compare}(b), differ from conventional MIL frameworks~\cite{ABMIL,CLAM,transmil,dsmil,dtfdmil}, as shown in Fig.~\ref{intro_compare}(a), by incorporating linguistic supervision alongside visual understanding. 
Most VLFM-based few-shot WSI methods~\cite{TOP,FiVE,PEMP,MSCPT,FAST,MIC24,li2024few} address few-shot classification by introducing various prompt-tuning techniques. For example, TOP~\cite{TOP} leverages pathology prior knowledge in its prompting, FiVE~\cite{FiVE} integrates pathological reports into prompts, PEMP~\cite{PEMP} enhances prompts with additional image references, and FAST~\cite{FAST} employs supplementary visual prompts through cached samples. 
% However, these methods mainly focus on prompt engineering at the input side, with limited improvements in the classifier design. [no improvement, use widely used, comprising xxx]
However, these methods mainly focus on prompt engineering at the input side while adopting the conventional classifier comprising a learnable aggregator (e.g., attention pooling) to generate global representations and a visual-linguistic similarity matching for predictions.
Our experimental analysis reveals this design's vulnerability for few-shot learning, with TOP~\cite{TOP} exhibiting a 25\% AUC drop (0.79$\to$0.54) by decreasing the training samples from 16 to 2 (see Tab.~\ref{results_nsclc}). Such performance degradation is primarily due to the overfitting of parameter-intensive aggregators under data scarcity.

% generally comprise parameter-intensive structure, which is prone to overfit with a few data. 

% As results shown in Tab.~\ref{results_nsclc}, xxx  exhibit 20\% performance degradation when trained with 8-shot \xmli{revise.} This is because these mainly rely on xxx learnable xxx fc to fine-tune xx, however, data is xxx, xxx. 

% Inspired by this, Notably, our baseline simply uses x to achieve x
% [1 refer method, 2. key motivation]
In light of this, we consider using non-parametric operations as an alternative to the attention-based aggregator.
% Initial experiments demonstrate that non-parametric top-K similarity matching achieves promising performance. As shown in Tab.~\ref{results_nsclc}, this baseline surpasses state-of-the-art prompt-tuning methods (CoOp \cite{CoOp} and TOP \cite{TOP}) by at least 4.9\% in AUC across various few-shot settings.
Notably, our baseline uses non-parametric top-K similarity matching to achieve promising performance, surpassing state-of-the-art prompt-tuning methods (CoOp \cite{CoOp} and TOP \cite{TOP}) by at least $4.9\%$ in AUC across various few-shot settings (see Tab.~\ref{results_nsclc}). 
This finding highlights the critical role of classifier design, i.e., cosine similarity, in few-shot scenarios.
% However, as top-K similarity matching is only one potential design for the classifier, relying solely on prioritizing patches with maximal disease descriptor alignment may result in suboptimal outcomes. 
However, relying solely on top-K similarity matching, which prioritizes patches with maximal disease descriptor alignment, may yield suboptimal outcomes.
Therefore, we further propose a meta-learning-based method that jointly leverages multiple classifiers with complementary diagnostic emphases, enabling a more effective identification of the most representative patches from the slide for diagnosis.
% enabling a more comprehensive and representative patch selection for accurate and robust diagnosis.
% Our key motivation is that by jointly leveraging multiple classifiers with complementary diagnostic emphases, we can more effectively identify the most representative patches from the slide for diagnosis.

% ??While the top-k similarity matching is one feasible design for the classifier by prioritizing patches with maximal disease descriptor alignment, this approach may suboptimally address tasks requiring discriminative patch characterization.??
% If we jointly utilize multiple classifiers with complementary diagnostic emphases, we could measure the patch more comprehensively.

%  that functions as a router to select and ensemble predictions from multiple classifiers
% (e.g., xxx, xxx, xxx, xxx)
% constructs optimal classifier schemes tailored to each individual input, 
In this work, we introduce a novel Meta-Optimized Classifier (MOC) for few-shot whole slide image classification. MOC comprises two core components: (1) a meta-learner that automatically optimizes a classifier configuration from a mixture of candidate classifiers and (2) a classifier bank housing diverse candidate classifiers to enable a holistic pathological interpretation. Specifically, the classifier bank is empirically implemented with four non-parametric operations, emphasizing three key aspects: maximal similarity, categorical prominence, and minimal irrelevance. 
Extensive experiments demonstrate that MOC consistently outperforms prior art across multiple few-shot benchmarks. Notably, on the TCGA-NSCLC benchmark, MOC achieves a significant improvement of $10.4\%$ in AUC over the state-of-the-art few-shot methods, with performance gains amplifying to $26.25\%$ under 1-shot conditions. This advancement is particularly critical for clinical deployments, where diagnostic training data is severely limited.

\section{Methodology}
% \subsection{Overview and Problem Formulation}

\begin{figure}[htb]
\centering
\includegraphics[width=\textwidth]{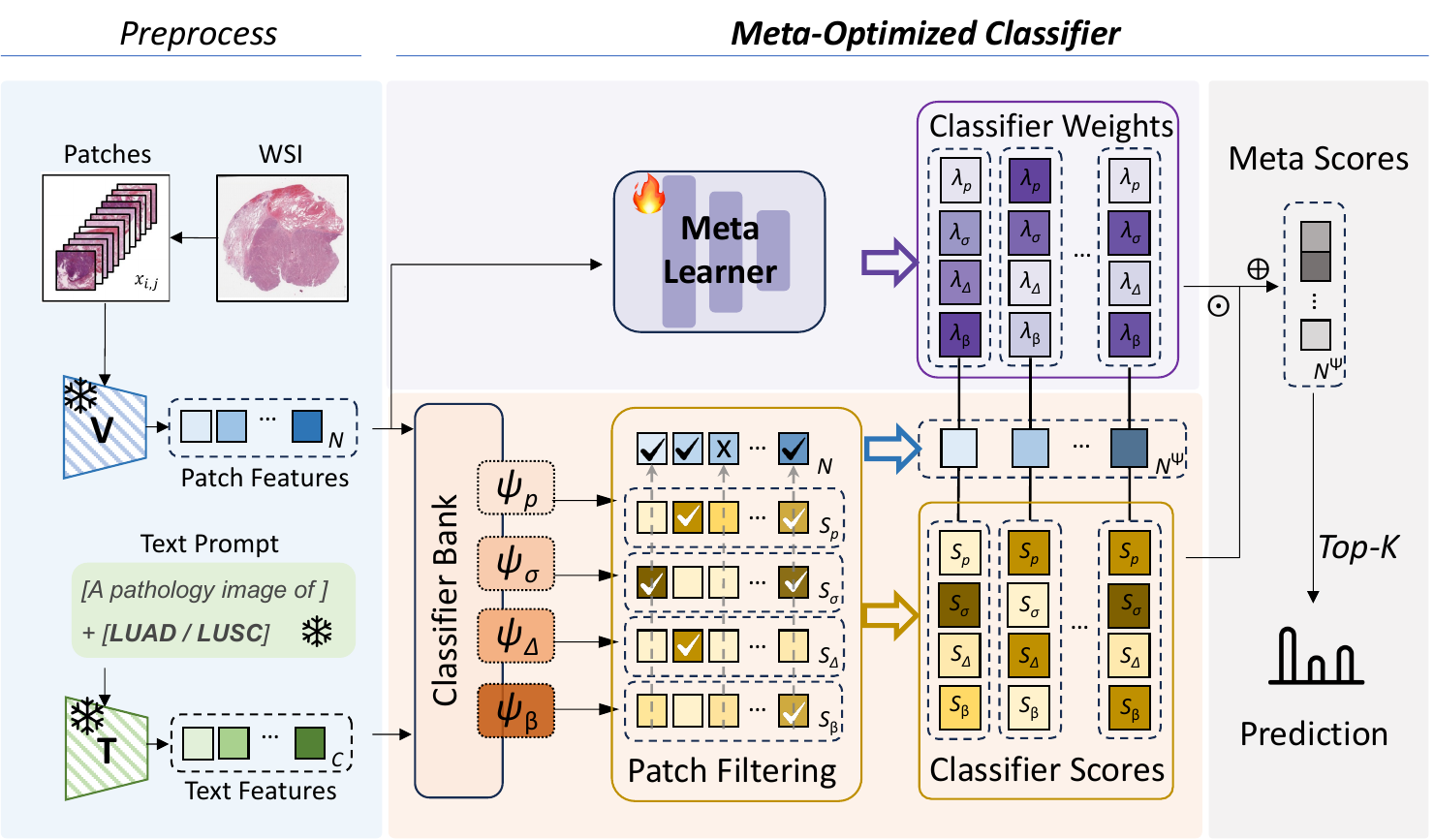}
% \vspace{-7mm}
\caption{The two-phase few-shot WSI classification pipeline: 1) Preprocess WSI and prompts. 2) WSI prediction with the proposed Meta-Optimized Classifier.} 
\label{pipeline}
\end{figure}

In this section, we introduce a novel few-shot WSI classification framework centered around our Meta-Optimized Classifier (MOC). As illustrated in Fig.~\ref{pipeline}, the framework begins with WSI preprocessing, followed by the application of MOC, which leverages a meta-learner to dynamically propose configurations and integrate candidate classifiers from the classifier bank. We will first elaborate on how the MOC assists few-shot WSI classification, followed by our construction recipe for the classifier bank.

The problem definition of our few-shot WSI classification task is briefly presented as follows: given a dataset $\mathcal{D} = \{X_1, X_2, ..., X_N\}$ comprising $N$ WSIs with $C$ distinct categories, each WSI $X_i$ is given a label $Y_i\in \{1,2,...,C\}$. Each WSI $X_i$ is then partitioned into $n_i$ non-overlapping patches $\{x_{i,j}, j = 1,2,...,n_i\}$, where the label for each patch $x_{i,j}$ is unknown. "Shot" refers to the number of labeled WSIs for each category, i.e., the training set for $C$-Category $K$-Shot WSI classification contains $K\times C$ labeled WSIs. Typically, $K$ can take values such as 1, 2, 4, or 8.

% \subsection{Patch Scoring and Filtering with Policy Bank}
\subsection{Meta-Optimized Classifier for Few-Shot WSI Classification}
The core function of the MOC is to construct an optimal classifier for each input instance. This is accomplished by the collaboration of the meta-learner and the classifier bank. Candidates in the classifier banks provide pathological analysis from different aspects, while the meta-learner coordinates such observations for an optimal classifier scheme.

%Therefore, we have the prompt set $T =\{t_c\}_{c = 1,...,C}$ for all $C$ categories. 

To be specific, we firstly \textit{preprocess} the WSIs. We utilize the pathology vision-language foundation model as the backbone of the proposed framework, which consists of pre-trained visual encoder $\mathcal{F(\cdot)}$ and text encoder $\mathcal{G(\cdot)}$. 
For each category $c$, the prompt $t_c$ = template + $\mbox{classname}_c$, where the template in the form of \texttt{`A pathology image of \{\}'} and the $\mbox{classname}_c$ takes this full name of the category, i.e., \texttt{`lung adenocarcinoma'}. Then, the corresponding prompt embeddings can be computed via $w_c = \frac{\mathcal{G}(t_c)}{\|\mathcal{G}(t_c)\|},$ and the $l_2$-normalized WSI embeddings for patch $x_{i,j}$ is given by $u_{i,j} = \frac{\mathcal{F}(x_{i,j})}{\|\mathcal{F}(x_{i,j})\|}.$ Therefore, we obtain the prompt embeddings set $W = \{w_c\}_{c=1,...,C}$ for all categories.

%and the WSI embeddings set $U_i=\{u_{i,j}\}_{j= 1,...,n_i}$ for each WSI $X_i$.
%\begin{equation}
 %  u_j = \frac{\mathcal{F}(x_j)}{\|\mathcal{F}(x_j)\|},\quad w_i = \frac{\mathcal{G}(t_i)}{\|\mathcal{G}(t_i)\|}.
%\end{equation}

%Each candidate $\psi_h$ is capable of taking an patch embedding $u_{i,j}$ for any patch $x_{i,j} $and a set of prompt embeddings $W$, then returning $S=\psi_h(u,W)$ as the image score. 

Defined the classifier bank as a collection of $H$ candidate classifiers: $\Psi = \{\psi_1, \psi_2, ..., \psi_H\}$. Here, each candidate classifier $\psi_h$ is capable of mapping patch embedding $u_{i,j}$ and the prompt embeddings set $W$ to a patch score $S^{\Psi_h}_{x_{i,j}}=\psi_h(u_{i,j},W)$. We demonstrate an effective detailed implementation for the classifier bank in Sec.~\ref{bank_sec}. The patch scores are subsequently utilized for patch filtering. 
% For instance, CLIP~\cite{CLIP} adopt a direct vision-language similarity as the classifier $\psi_{\mbox{clip}}$, where given the image embedding $u$ and the text embeddings $W$, the score for this image is calculated via $S = \psi_{\mbox{clip}}(u,W) = u^TW$.
% The classifier bank can have various implementations, while more diversified candidates generally provide a more holistic understanding, which consequently leads to better outcomes. We demonstrate an effective implementation in Sec.~\ref{bank_sec}.
%With the classifier bank, we then perform patch scoring and filtering. 
%For a given patch $x_j$ with embedding $u_j$ and the prompt embeddings $W$, we obtain the classifier bank-based score set $S_{x_j}^{\Psi}=\{S_{x_j}^{\psi_h}\}_{h=1,...,H}$, each computed via:
%\begin{equation}\label{patch_score}
 %   S_{x_j}^{\psi_h} = \psi_h(u_j, W).
%\end{equation}
Specifically, each candidate classifier $\psi_h$ elects a subset of patches for WSI $X_i$, denoted as $Bag_{X_i}^{\psi_h}$, which includes the top $q$ patches with the highest scores:
% Specifically, each candidate classifier $\psi_h$ will nominate a subset patches for WSI $X$, denoted as $Bag_X^{\psi_h}$, containing $q$ patches with the highest score:
% where we further create the representative bag $Bag_X^\Psi$ for the WSI $X$ via:
\begin{equation}\label{eq_bag}
    Bag^{\psi_h}_{X_i} = \{x_{i,j} \ |\ x_{i,j} \in \mathop{argmax}\limits^{(q)} S_{x_{i,j}}^{\psi_h}\}, \ \forall\psi_h \in \Psi.
\end{equation}
Subsequently, a bank-nominated set $Bag_{X_i}^\Psi$ for the WSI $X_i$ is obtained by taking a union of these sets: $Bag_{X_i}^\Psi = \mathop\bigcup\limits_{\psi_h \in \Psi}Bag_{X_i}^{\psi_h}$. This filtering process removes patches with limited significance by a consensus among all candidate classifiers.

Further, the meta-learner 
$\mathcal{M}$, structured as a two-layer perceptron, predicts the classifier weights for the $H$ candidate classifiers based on a patch embedding $u_{i,j}$. Specifically, given a nominated patch $x_{i,j} \in Bag_{X_i}^\Psi$ for WSI $X_i$, we obtain a set of $H$ classifier weights $\Lambda_{{i,j}}=\{\lambda^1_{x_{i,j}},\lambda^2_{x_{i,j}},..., \lambda^H_{x_{i,j}}\}$ 
 by:
\begin{equation}
    \Lambda_{{i,j}} = \mathcal{M}(u_{i,j}).
\end{equation} And the set of nominated patch prediction $p_{X_i} = \{p_{x_{i,j}}\}_{x_{i,j} \in Bag_{X_i}^\Phi}$ is calculated via:
\begin{equation}
    p_{x_{i,j}} = \sum_{h=1}^H \lambda^h_{x_{i,j}} \cdot S_{x_{i,j}}^{\psi_h}.
\end{equation}

Finally, we use a top-K max-pooling operation, $h_{\mbox{topK}}$, to get the WSI-level prediction $\mathcal{P}_{X_i}$ for WSI $X_i$:
\begin{equation}\label{eq_topk}
    \mathcal{P}_{X_i} = h_{\mbox{topK}}(p_{X_i}) = \frac{1}{K} \left [ \sum\limits_{i=1}^K \Tilde{p}_i^1, \sum\limits_{i=1}^K \Tilde{p}_i^2, ..., \sum\limits_{i=1}^K \Tilde{p}_i^C \right ],
\end{equation}
where $\Tilde{p}_i^c$ is the $i\mbox{-}th$ largest score values from patch-level prediction set $p_{X_i}$ for category $c$. The parameter of the meta-learner is optimized with the cross-entropy loss between the WSI-level prediction and ground truth.

\subsection{Classifier Bank Construction with diverse Classifiers}\label{bank_sec}
Candidate classifiers in the classifier bank aim to provide complementary diagnostic emphases, allowing the meta-learner to comprehensively recognize the significance of each patch. 
% Therefore, more diversified candidates generally provide a more holistic understanding, which consequently leads to a better outcome. 
In this section, we demonstrate an effective way to construct the classifier bank with four candidate classifiers, denoted as $\Psi = \{\psi_p, \psi_\sigma, \psi_\Delta, \psi_\beta\}$. Each candidate classifier is illustrated as follows:

\begin{itemize}
    \item[{$\bullet$}] \textbf{Confidence Peak Classifier} $\psi_p$ evaluates maximal similarity by computing the cosine similarity between the patch embedding and each prompt embedding, formulated as
    \begin{align}
        S_{x_{i,j}}^{\psi_p} = u_{i,j}^TW.
    \end{align}

    \item [{$\bullet$}] \noindent\textbf{Normalized Certainty Classifier} $\psi_\sigma$ identifies easy distinguishability by applying a softmax function $\sigma$ to the cosine similarity between patch embeddings and prompt embeddings. This is defined as:
    \begin{align}
        S_{x_{i,j}}^{\psi_\sigma} = \sigma(u_{i,j}^TW).
    \end{align}

    \item [{$\bullet$}] 
    \noindent\textbf{Divergence Extremum Classifier} $\psi_\Delta$ also evaluates easy discrimination by computing the similarity difference of the highest two categories, formulated as:
    \begin{align}
        S_{x_{i,j}}^{\psi_\Delta} = max_{(1)}(u_{i,j}^TW) - max_{(2)}(u_{i,j}^TW).
    \end{align}

    \item [{$\bullet$}] \noindent\textbf{Background Suppression Classifier} $\psi_\beta$ measures the minimal irrelevance by calculating the negative similarity between a patch and the background tissues. Specifically, we additionally introduce four normal tissue types, i.e., \texttt{`stromal tissue'}, \texttt{`inflammatory tissue'}, \texttt{`vascular tissue'}, and \\\texttt{`necrotic tissue'}. Following the same process routine as foreground categories, we obtain the background prompts $\{t^\beta_c\}_{c=1,...,C^\beta}$ and background prompt embedding $W^\beta = \{w^\beta_c\}_{c=1,...,C^\beta}$, where $C^\beta$ denotes the number of background tissue types. Then, the patch score is given by:
    \begin{align}
        S_{x_{i,j}}^{\psi_\beta} = -\sum\limits_{c=1}^{C^\beta} u^T_{i,j}w_c^\beta.
    \end{align}
    
\end{itemize}

\section{Experiments}
\noindent\textbf{Dataset.}
We comprehensively compare our proposed MOC with many state-of-the-art (SoTA) methods using two real-world datasets: TCGA-NSCLC and TCGA-RCC. TCGA-NSCLC consists of 1052 WSI slides for lung cancer subtyping, and TCGA-RCC consists of 937 WSI slides for kidney cancer subtyping.
% TCGA-NSCLC consists of 540 slides of lung adenocarcinoma (LUAD) and 512 slides of lung squamous cell carcinoma (LUSC). TCGA-RCC consists of 121 slides of kidney chromophobe carcinoma (KICH), 519 slides of kidney clear cell carcinoma (KIRC), and 297 slides of kidney papillary carcinoma (KIRP). 
Each dataset is randomly split five times into the training, validation, and test sets. For TCGA-NSCLC, each fold consists of 50 validation samples and 200 test samples, whereas for TCGA-RCC, each fold contains 20 validation samples and 70 test samples. To simulate the few-shot learning scenario, we then randomly select $k$ samples per category from the training set to construct the $k$-shot experimental setup.

\begin{table}[!b]
\begin{centering}
\caption{Few-shot results on TCGA-NSCLC dataset. The best results are in bold, and the second-best results are underlined.}
\resizebox{1\textwidth}{!}{
\setlength{\tabcolsep}{2pt}
\label{results_nsclc}
\begin{tabular}{l|cc|cc|cc|cc}
\toprule
\multirow{2}{*}{Method} &
  \multicolumn{2}{c|}{1 shot} &
  \multicolumn{2}{c|}{2 shot} &
  \multicolumn{2}{c|}{4 shot} &
  \multicolumn{2}{c}{8 shot} \\ \cline{2-9} 
                            & AUC (\%)   & Acc. (\%)   & AUC (\%)    & Acc. (\%)   & AUC (\%)    & Acc. (\%)   & AUC (\%)    & Acc. (\%)   \\ \midrule
% ABMIL    & 52.39 & 48.38 & 58.03 & 48.23 & 69.22 & 51.95 & 73.68 & 59.96 \\
% \hline 
\multicolumn{9}{c}{MIL-based Methods}  \\ 
\midrule 
CLAM-SB~\cite{CLAM}         & $52.39_{\pm6.24}$ & $48.38_{\pm1.43}$ & $58.03_{\pm5.76}$ & $48.23_{\pm1.36}$ & $69.22_{\pm4.74}$ & $51.95_{\pm5.15}$ & $73.68_{\pm6.72}$ & $59.96_{\pm9.44}$ \\
CLAM-MB~\cite{CLAM}         & $58.75_{\pm9.81}$ & $50.57_{\pm2.78}$ & $65.40_{\pm4.41}$ & $61.87_{\pm2.68}$ & $71.92_{\pm4.14}$ & $52.43_{\pm3.15}$ & $79.69_{\pm6.34}$ & $66.63_{\pm2.48}$ \\
TransMIL~\cite{transmil}    & $62.24_{\pm4.59}$ & $55.24_{\pm5.15}$ & $63.95_{\pm4.95}$ & $55.87_{\pm5.17}$ & $74.51_{\pm5.66}$ & $61.42_{\pm7.95}$ & $83.69_{\pm7.05}$ & $\underline{75.82_{\pm7.44}}$ \\
ViLa-MIL~\cite{vilamil}     & $71.79_{\pm4.64}$ & $56.48_{\pm6.79}$ & $72.93_{\pm3.35}$ & $60.34_{\pm3.60}$ & $77.79_{\pm4.88}$ & $57.86_{\pm7.66}$ & $84.20_{\pm7.09}$ & $73.51_{\pm9.22}$ \\ 

\midrule 
 \multicolumn{9}{c}{Few-shot VLFM-based Methods}  \\ 
\midrule 
CoOp~\cite{CoOp}             & $62.04_{\pm8.24}$ & $59.15_{\pm7.84}$ & $70.21_{\pm2.78}$ & $63.97_{\pm2.47}$ & $72.23_{\pm4.18}$ & $58.65_{\pm3.42}$ & $80.11_{\pm5.91}$ & $69.48_{\pm5.93}$ \\
TOP~\cite{TOP}               & $54.76_{\pm6.95}$ & $51.23_{\pm4.09}$ & $65.79_{\pm2.87}$ & $57.48_{\pm3.33}$ & $72.67_{\pm9.00}$ & $65.90_{\pm7.51}$ & $79.43_{\pm8.30}$ & $71.02_{\pm7.52}$ \\
\midrule
$\dag$\textbf{MOC w/o. ($\mathcal{M}$,$\Psi$)}& \underline{$85.00_{\pm1.63}$} & $\mathbf{75.31_{\pm1.93}}$ & \underline{$85.00_{\pm1.63}$} & $\mathbf{75.31_{\pm1.93}}$ & \underline{$85.00_{\pm1.63}$} & $\mathbf{75.31_{\pm1.93}}$ & \underline{$85.00_{\pm1.63}$} & $75.31_{\pm1.93}$ \\
\rowcolor{pink!30} \textbf{Our MOC}   & $\mathbf{88.29_{\pm2.65}}$ & $\underline{73.95_{\pm3.73}}$ & $\mathbf{89.11_{\pm1.13}}$ & $\underline{74.26_{\pm5.21}}$ & $\mathbf{90.65_{\pm1.02}}$ & $\underline{68.57_{\pm3.80}}$ & $\mathbf{90.51_{\pm1.74}}$ & $\mathbf{77.10_{\pm3.80}}$ \\
\bottomrule
\end{tabular}}
\begin{threeparttable}
 \begin{tablenotes}
        \scriptsize
        \item[$\dag$] Report the average zero-shot results on all test sets.
\end{tablenotes}
\end{threeparttable}
\par\end{centering}
\end{table}

\noindent\textbf{Implementation and evaluation}
% backbone, preprocess, hyperparam, prompt
We employ the CONCH~\cite{CONCH} pretraining as the backbone for both the image and text encoders. We use the CLAM~\cite{CLAM} toolkit for WSI preprocessing, and we set the patch size to 224 for feature extraction. We use the same prompt ensemble scheme following MI-Zero~\cite{MI-Zero}. MOC w/o. ($\mathcal{M}$, $\Psi$) in Tab.~\ref{results_nsclc} and Tab.~\ref{results_rcc} is implemented following MI-Zero~\cite{MI-Zero}. For a fair comparison, we apply the same splits, backbone, preprocessing, and prompts to all the methods. The learning rate is set to $1e^{-3}$, $q$ in Eq.~\ref{eq_bag} is set to 1000, and $K$ in Eq.~\ref{eq_topk} is set to 150. 
Hyperparameters for baseline methods are following the original implementation. We respectively report the average Area Under the Curve (AUC) and Accuracy (ACC) with corresponding standard deviation ($\pm$).

\noindent\textbf{Remarkable improvements in few-shot WSI classification.} We compare the performance of our proposed MOC with many SoTA MIL-based methods and few-shot VLFM-based methods on the TCGA-NSCLC dataset as Tab.~\ref{results_nsclc} and the TCGA-RCC dataset as Tab.~\ref{results_rcc}. The results suggest our method achieves the best on both datasets among different few-shot settings. 
Specifically, on the TCGA-NSCLC dataset, our MOC surpasses the second-best performing few-shot WSI classification method by $26.25\%$, $18.9\%$, $17.98\%$, $10.4\%$ in AUC for 1, 2, 4, 8-shot settings and also outperforms zero-shot baseline by at least $3.29\%$.
% Note that we did not compare our method with FiVE~\cite{FiVE}, PEMP~\cite{PEMP}, and FAST~\cite{FAST} as they require additional annotations.

\begin{table}[tb]
\begin{centering}
\caption{Few-shot results on TCGA-RCC dataset. The best results are in bold, and the second-best results are underlined.}
\resizebox{1\textwidth}{!}{
\setlength{\tabcolsep}{1pt}
\label{results_rcc}
\begin{tabular}{l|cc|cc|cc|cc}
\toprule
\multirow{2}{*}{Method} &
  \multicolumn{2}{c|}{1 shot} &
  \multicolumn{2}{c|}{2 shot} &
  \multicolumn{2}{c|}{4 shot} &
  \multicolumn{2}{c}{8 shot} \\ \cline{2-9} 
                            & AUC (\%)   & Acc. (\%)   & AUC (\%)    & Acc. (\%)   & AUC (\%)    & Acc. (\%)   & AUC (\%)    & Acc. (\%)   \\ \midrule
% ABMIL    & 52.39 & 48.38 & 58.03 & 48.23 & 69.22 & 51.95 & 73.68 & 59.96 \\
\multicolumn{9}{c}{MIL-based Methods}  \\ 
\midrule
CLAM-SB~\cite{CLAM}         & $74.05_{\pm10.56}$ & $46.03_{\pm6.07}$ & $77.97_{\pm12.92}$ & $48.68_{\pm7.12}$ & $90.77_{\pm1.24}$ & $66.98_{\pm4.62}$ & $95.20_{\pm1.09}$ & $83.89_{\pm2.78}$ \\
CLAM-MB~\cite{CLAM}         & $75.81_{\pm13.37}$ & $56.49_{\pm14.22}$ & $77.97_{\pm14.35}$ & $44.09_{\pm5.91}$ & $91.34_{\pm2.45}$ & $76.68_{\pm5.16}$ & $95.53_{\pm1.30}$ & $82.35_{\pm3.39}$ \\
TransMIL~\cite{transmil}    & $81.49_{\pm10.36}$ & $46.25_{\pm7.82}$ & $84.17_{\pm9.52}$ & $63.05_{\pm14.31}$ & $93.50_{\pm1.88}$ & $80.21_{\pm3.73}$ & $96.07_{\pm1.59}$ & $80.77_{\pm8.06}$ \\
ViLa-MIL~\cite{vilamil}     & $82.55_{\pm12.31}$ & $59.33_{\pm12.75}$ & $82.92_{\pm10.43}$ & $63.83_{\pm11.95}$ & $93.15_{\pm2.64}$ & $79.02_{\pm5.97}$ & $95.61_{\pm2.36}$ & $\underline{84.07_{\pm5.37}}$ \\ 
\midrule
 \multicolumn{9}{c}{Few-shot VLFM-based Methods}  \\ 
\midrule 
CoOp~\cite{CoOp}            & $87.40_{\pm2.51}$ & $54.33_{\pm3.39}$ & $82.59_{\pm15.14}$ & $50.65_{\pm9.37}$ & $92.92_{\pm1.85}$ & $58.28_{\pm3.39}$ & $94.98_{\pm1.77}$ & $58.81_{\pm3.61}$ \\
TOP~\cite{TOP}              & $67.60_{\pm3.45}$ & $34.70_{\pm4.46}$ & $75.73_{\pm3.96}$ & $43.85_{\pm10.04}$ & $72.39_{\pm3.64}$ & $47.64_{\pm8.44}$ & $72.02_{\pm4.40}$ & $51.44_{\pm10.23}$ \\ \midrule
$\dag$\textbf{MOC w/o. ($\mathcal{M}$, $\Psi$)} & $\underline{96.03_{\pm0.68}}$ & $\underline{80.39_{\pm2.33}}$ & $\underline{96.03_{\pm0.68}}$ & $\underline{80.39_{\pm2.33}}$ & $\underline{96.03_{\pm0.68}}$ & $\underline{80.39_{\pm2.33}}$ & $\underline{96.03_{\pm0.68}}$ & $80.39_{\pm2.33}$ \\
\rowcolor{pink!30} \textbf{Our MOC}    & $\mathbf{96.25_{\pm1.41}}$ & $\mathbf{84.34_{\pm5.24}}$ & $\mathbf{97.45_{\pm0.72}}$ & $\mathbf{90.02_{\pm2.55}}$ & $\mathbf{97.42_{\pm0.41}}$ & $\mathbf{89.39_{\pm2.75}}$ & $\mathbf{97.78_{\pm0.40}}$ & $\mathbf{91.74_{\pm1.44}}$ \\
\bottomrule
\end{tabular}} 
\begin{threeparttable}
 \begin{tablenotes}
        \scriptsize
        \item[$\dag$] Report the average zero-shot results on all test sets.
\end{tablenotes}
\end{threeparttable}
\par\end{centering}
\end{table}

\noindent\textbf{Ablation study}
% As shown in Tab.~\ref{ablation1}, our ablation results reveal a positive correlation between candidate classifiers' diversity and model performance. Specifically, Tab.~\ref{ablation1}(a) presents the increasing improvements achieved through successive integration $\psi_\beta$, $\psi_\Delta$, and $\psi_\sigma$ into baseline $\psi_p$. Furthermore, Tab.~\ref{ablation1}(b) analyzes all possible classifier combinations. Ablation demonstrates that an increasing number of classifiers gives better performance, underscoring the critical role of a diverse classifier bank in enhancing model effectiveness.
As shown in Tab.~\ref{ablation1}(left), our proposed MOC achieves the highest performance (89.64\%) compared to the other three methods, demonstrating incremental performance gains through the successive integration $\psi_\beta$, $\psi_\Delta$, and $\psi_\sigma$ into baseline $\psi_p$. Tab.~\ref{ablation1}(right) reveals that the method incorporating four classifiers achieves the highest performance among all configurations, highlighting a consistent performance improvement as the number of classifiers increases.

% As shown in Table~\ref{ablation1}, our ablation study highlights a strong positive correlation between the diversity of candidate classifiers and model performance.
% Table~\ref{ablation1}(a) demonstrates incremental performance gains via successive integration $\psi_\beta$, $\psi_\Delta$, and $\psi_\sigma$ into baseline $\psi_p$.
% Table~\ref{ablation1}(b) further analyzes all possible classifier combinations, showing the performance gain with an increasing number of classifiers. These results underscore the importance of a diverse classifier bank in improving model effectiveness.
% demonstrating the effectiveness of incorporating a diverse classifier bank.
% demonstrating that ensembles containing more classifiers consistently outperform simpler combinations, with our four-classifier implementation achieving the highest performance.

\begin{table}[!h]
\centering
\caption{Comparison of methods with (left) varying classifier combinations and (right) increasing number of classifiers (considering all possible classifier combinations).}
% \caption{Ablation study on the impact of classifier diversity. (a) Incremental performance gains from successive integration of classifiers, where x and y denote incomplete implementation of our proposed MOC. (b) Performance gain with an increasing number of classifiers considering all possible classifier combinations.}
\label{ablation1}
\begin{minipage}[t]{0.58\textwidth}
\setlength{\tabcolsep}{3pt}
\begin{tabular}{l|c|c|c|c|c}
\hline
Method                  & $\psi_p$ & $\psi_\sigma$  & $\psi_\Delta$ & $\psi_\beta$ & avg AUC(\%) \\ \hline
MOC w/o. ($\mathcal{M}$, $\Psi$)  & \ding{52} & \ding{55} & \ding{55} & \ding{55} & 85.00          \\ 
Method x                  & \ding{52} & \ding{55} & \ding{55} & \ding{52} & 87.63           \\ 
Method y                  & \ding{52} & \ding{55} & \ding{52} & \ding{52} & 88.95           \\
\rowcolor{pink!30}\textbf{MOC}   & \ding{52} & \ding{52} & \ding{52} & \ding{52} & 89.64           \\ \hline
\end{tabular}
% \vspace{-8pt}
% \caption*{(a)}
\end{minipage}
\begin{minipage}[t]{0.4\textwidth}
\centering
\setlength{\tabcolsep}{3pt}
\begin{tabular}{c|c}
\hline
\#Classifiers           & avg AUC(\%) \\ \hline
1                       & 85.00           \\ 
2                       & 86.59           \\
3                       & 88.82           \\
\rowcolor{pink!30}4     & 89.64           \\
\hline
\end{tabular}
% \vspace{-8pt}
% \caption*{(b)}
\end{minipage}
\end{table}

The comparative analysis in Tab~\ref{ablation2} provides critical insights into classifier integration strategies. Given multiple classifiers, a naive summation yields only marginal improvements (merely 0.4\% increase over baseline), while our meta-learner-based integration achieves a remarkable 4.64\% AUC enhancement, effectively leveraging complementary classifiers for a more holistic understanding.
% This performance gap empirically validates our hypothesis that a learned scheme providing dynamic weights significantly outperforms static fusion approaches.

\begin{table}[!tb]
\setlength{\tabcolsep}{7pt}
\centering
% \caption{Effectiveness of meta-learner. Report average results of 1,2,4,8-shot.}
% \caption{Ablation results of meta-learner: Meta-learned integration surpasses static fusion. Report average AUC of 1, 2, 4, 8-shot experiments.}
\caption{Performance comparison of our baseline, multiple classifiers fused with summation, and our proposed MOC.}
\label{ablation2}
\begin{tabular}{l|c}
\hline
Method                                                  & average AUC(\%)     \\ \hline
MOC w/o. ($\mathcal{M}$, $\Psi$)                                            & 85.00   \\ 
Multiple Classifiers (w. Summation)                     & 85.40   \\ 
\rowcolor{pink!30} \textbf{MOC (w. Meta-learner)}       & 89.64   \\
\hline
\end{tabular}
\end{table}

\begin{figure}[!htb]
\centering
\includegraphics[width=1.0\textwidth]{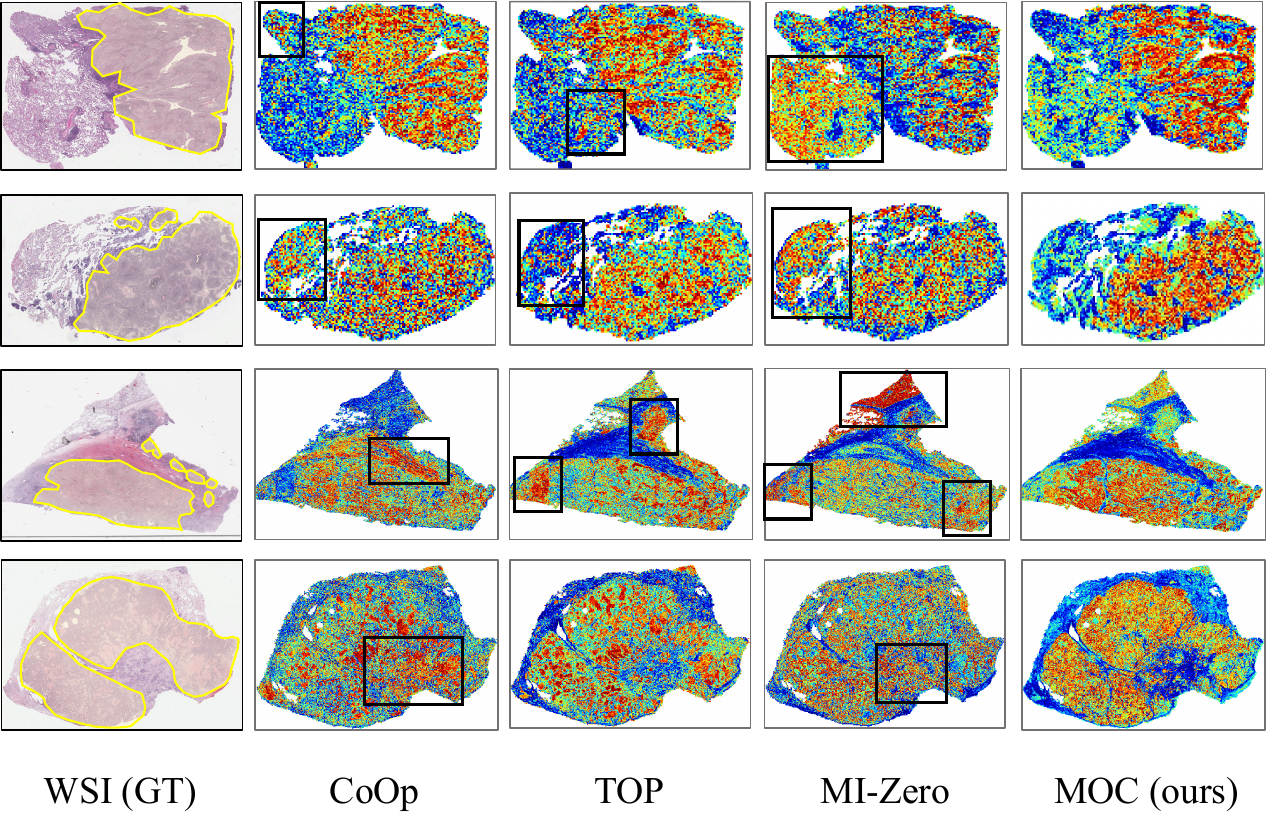}
\caption{Visualizations on the TCGA-NSCLC datasets. Ground-truth tumors are circled in yellow,  predicted tumor regions are red, and black boxes note false positives.}
\label{fig_vis}
\end{figure}

\noindent\textbf{Qualitative visualization}
We further depict the visualizations in Fig. \ref{fig_vis}. We compare our method with CoOp~\cite{CoOp}, TOP~\cite{TOP}, and MI-Zero~\cite{MI-Zero}. We find our MOC obviously shows better results than other baselines. 
% Specifically, the baseline methods predict more false positives.

\section{Conclusion}
% In summary, this study presents a novel and effective Meta-Optimized Classifier (MOC) for few-shot WSI classification that dynamically integrates diverse non-parametric classifiers. To address the limitations of the conventional classifiers widely adopted by existing few-shot VLFM-based
% Extensive experiments and ablations have shown the state-of-the-art performance of MOC and the effectiveness of our design.

In summary, this study presents a novel Meta-Optimized Classifier (MOC) for few-shot WSI classification. To address the limitation of the classifier design in existing VLFM-based few-shot explorations, we innovatively use a meta-learner to dynamically construct an optimal classifier from the classifier bank. Furthermore, we implement the classifier bank with diverse classifiers for a holistic pathological understanding. Extensive experiments demonstrate our MOC's state-of-the-art performance among multiple few-shot benchmarks.

\begin{credits}
\subsubsection{\ackname}
This work is supported by a research grant from the National Natural Science Foundation of China (Grant No. 62306254) and a research grant from the Research Grants Council of the Hong Kong Special Administrative Region, China (Project No. R6005-24).

\subsubsection{\discintname}
The authors have no competing interests to declare that are relevant to the content of this article.
\end{credits}

\bibliographystyle{splncs04}
\bibliography{Paper-0042}

\end{document}